\documentclass[conference]{IEEEtran}
\IEEEoverridecommandlockouts
\usepackage{cite}
\usepackage{amsmath,amssymb,amsfonts}
\usepackage{graphicx}
\usepackage{textcomp}
\usepackage{xcolor}

\usepackage{graphics} 
\usepackage{enumerate}

\usepackage{algpseudocode}  
\usepackage{multirow}
\usepackage{verbatim}
\usepackage{url}
\newcommand{\tabincell}[2]{\begin{tabular}{@{}#1@{}}#2\end{tabular}}  

\def\BibTeX{{\rm B\kern-.05em{\sc i\kern-.025em b}\kern-.08em
    T\kern-.1667em\lower.7ex\hbox{E}\kern-.125emX}}
\begin{document}

\title{VGPN: Voice-Guided Pointing Robot Navigation for Humans
\thanks{Taijiang Mu is the corresponding author. {e-mail: mmmutj@gmail.com}.}
}

\author{\IEEEauthorblockN{Jun Hu, Zhongyu Jiang, Xionghao Ding, Taijiang Mu}
\IEEEauthorblockA{
\textit{Department of Computer Science and Technology}\\
\textit{Tsinghua University}\\
Beijing, China \\
huj17@mails.tsinghua.edu.cn, zyjiang@uw.edu, \\
dingxionghao14@gmail.com, mmmutj@gmail.com}
\and
\IEEEauthorblockN{Peter Hall}
\IEEEauthorblockA{\textit{Department of Computer Science} \\
\textit{University of Bath}\\
Bath, United Kingdom \\
P.M.Hall@bath.ac.uk}
}

\maketitle

\begin{abstract}
Pointing gestures are widely used in robot navigation approaches nowadays. However, most  approaches only use pointing gestures, and these have two major limitations. Firstly, they need to recognize pointing gestures all the time, which leads to long processing time and significant system overheads. Secondly, the user's pointing direction may not be very accurate, so the robot may go to an undesired place. To relieve these limitations, we propose a voice-guided pointing robot navigation approach named VGPN, and implement its prototype on a wheeled robot, TurtleBot 2. VGPN recognizes a pointing gesture only if voice information is insufficient for navigation. VGPN also uses voice information as a supplementary channel to help determine the target position of the user's pointing gesture. In the evaluation, we compare VGPN to the pointing-only navigation approach. The results show that VGPN effectively reduces the processing time cost when pointing gesture is unnecessary, and improves the user satisfaction with navigation accuracy.
\end{abstract}


\section{Introduction}
With the rapid development of robot technology, robots  are becoming ever more 
important. As an important research topic, robot navigation interaction aims 
to provide an intuitive and natural way for humans to control the movement 
of robots. 

Voice is a commonly used way of robot navigation for humans, and 
many previous approaches \cite{Lv2008Robot, tellex2011understanding, T2014Determining, Muthugala2016Enhancing, Muthugala2017Interpreting} are based on it. This method 
requires the user to describe the target place. But a significant 
limitation of this method occurs when the target place is hard to describe, in which case the 
user is forced to use extended descriptions comprising many words.
Additionally,  some words may be  hard for the robot to understand, such as uncertain terms and numbers (``go {\em near} to the {\em eleventh} chair behind the bed''). 

Compared to voice, using pointing gesture is a more intuitive and natural way, and recently more and more approaches \cite{Sato2007Natural, Yoshida2011Evaluation, Van2011Real, Abidi2013Human, T2017Foundations} are 
based on it. This way only requires the user to point at the target place 
without speaking at all; so this way is much more convenient. However,  pointing-only
  has two major limitations in practical use. Firstly,  pointing gesture recognition is often time-consuming because of complex image processing, and the pointing-only way needs to recognize pointing gestures continuously, which leads to long processing time and considerable system overheads. Secondly, the user's pointing direction may not be very accurate, so the robot may go to an undesired place.

To improve the performance of pointing-only robot navigation, we propose to use voice information to guide it. In this way, the user points at the target place and issues simple voice commands that contain demonstrative pronouns such as ``that'' and ``there''. By using voice 
guidance, pointing robot navigation can achieve better efficiency and accuracy.
For example, there are two common scenarios:

\begin{itemize}
\item {\em Scene 1.} Speech is sufficient for robot navigation, and 
pointing gesture is unnecessary. For example, when the user says ``go to that 
door'' and points at the door, if there is only one door nearby, the robot can 
directly go to the door, without needing to understand the user's pointing gesture. Another 
example is that, when the user says ``go forward'' and points forwards, the 
robot can just go forward, the pointing gesture need not be recognized.

\item {\em Scene 2.} The pointing direction is not very accurate, and  speech 
describes the navigation target. For example, if there are three chairs not so close to each other, when the user says ``go to 
that chair'' and points at the bed nearby the desired chair, the robot should go to the 
chair, because the user very probably means the chair.
\end{itemize}

In this paper, we propose a voice-guided pointing robot
navigation approach named VGPN, which can improve both the efficiency and the accuracy 
of pointing robot navigation. To improve efficiency, VGPN recognizes a pointing gesture 
only if voice information is insufficient for navigation. Consequently, when pointing gesture is unnecessary, VGPN 
can effectively reduce processing time costs and system overheads caused by recognizing pointing 
gesture continuously. To improve accuracy, VGPN uses voice information as 
supplementary information to help determine the target place of the user's pointing 
gesture. We implement the prototype of VGPN on a wheeled robot, TurtleBot 2. And by
using an existing SLAM approach \cite{Hess2016Real}, VGPN can navigate the robot to a 
target place unrestricted by sensor range.

We make two main contributions in this paper:
\begin{enumerate} [(1)]
\item We propose a voice-guided pointing robot navigation approach named VGPN, which 
can improve both the efficiency and accuracy of pointing robot navigation. 
\item We implement the prototype of VGPN on a wheeled robot, TurtleBot 2, and 
evaluate  its efficiency and accuracy. The results show that, compared to the pointing-only approach  without voice guidance, VGPN reduces 79.8\% processing time costs when pointing gesture is unnecessary, and also effectively 
improves user satisfaction with navigation accuracy.
\end{enumerate}

The remainder of this paper is organized as follows. Section II shows the 
related work of pointing-based robot navigation. Section III introduces VGPN and its 
implementation in detail. Section IV presents the evaluation of VGPN. Finally, 
Section V concludes this paper.

\section{RELATED WORK}

\subsection{Pointing-based Robot Navigation} 

Using pointing gestures is a natural way to interact with robots to specify the spatial position of a target location, and many 
approaches use it to perform robot navigation
\cite{Sato2007Natural, Yoshida2011Evaluation, Van2011Real, Abidi2013Human, 
T2017Foundations}. These approaches translate pointing gestures 
into a goal for the robot, and navigate the robot to that goal. Yoshida et al. 
\cite{Yoshida2011Evaluation} propose a pointing navigation approach based on a 
fish-eye camera. However, to use this approach, the user has to wear a bright and 
uniform color jacket and gloves. Moreover, the user has to keep pointing until the robot reaches the desired position in \cite{Yoshida2011Evaluation, Abidi2013Human}. The approach proposed in \cite{T2017Foundations} requires the 
user to raise their left hand to trigger the movement of the robot, so 
this way is not natural.

Existing pointing-only approaches for robot navigation have two major
limitations. Firstly, they need to recognize pointing gestures all the time, which is costly in time and resource.   Secondly, the accuracy of  robot navigation heavily depends on the accuracy of the user's pointing direction.

\subsection{Voice- and Pointing-based Human-Robot Interaction}
Combining voice and pointing gestures is an interesting way for Human-Robot 
Interaction(HRI), and it has received close attention. There are many HRI approaches 
\cite{Richard1980, Stiefelhagen2004Natural, carbini2006wizard, lei2014artificial, tscharn2017stop, Yan2017Task} that are based on this combination. Bolt 
\cite{Richard1980} proposes  a ``Put-That-There'' natural graphics interface using 
voice and pointing gestures. Stiefelhagen et al. \cite{Stiefelhagen2004Natural} 
present a robot for natural multi-modal HRI using speech, head pose and 
pointing gestures. Tscharn et al. \cite{tscharn2017stop} combine voice and pointing 
gestures to instruct a car about desired interventions which 
include spatial references. In \cite{Yan2017Task},  a 
robot can execute a vague task by combining verbal language and pointing 
gestures. For robot navigation, there are only few existing approaches that combine voice and 
pointing gestures. \cite{lei2014artificial} is such an approach, but it still needs to recognize pointing gestures all the time.

Combining voice and pointing gestures is a commonly used way in HRI, but it is hardly directly used in robot navigation.

\section{APPROACH}
In this section, we introduce VGPN and describe its implementation in detail.

Figure 1 shows the overall procedure of VGPN, which consists of three phrases:
\begin{itemize}
\item \textbf{Voice understanding.} VGPN receives a voice signal from the user and 
understands voice commands.
\item \textbf{Pointing direction estimation.} VGPN calculates the pointing 
direction of the user.
\item \textbf{Target decision.} VGPN decides the target position using voice 
guidance.
\end{itemize}

\begin{figure}[thpb]
  \centering
  \includegraphics[scale=1]{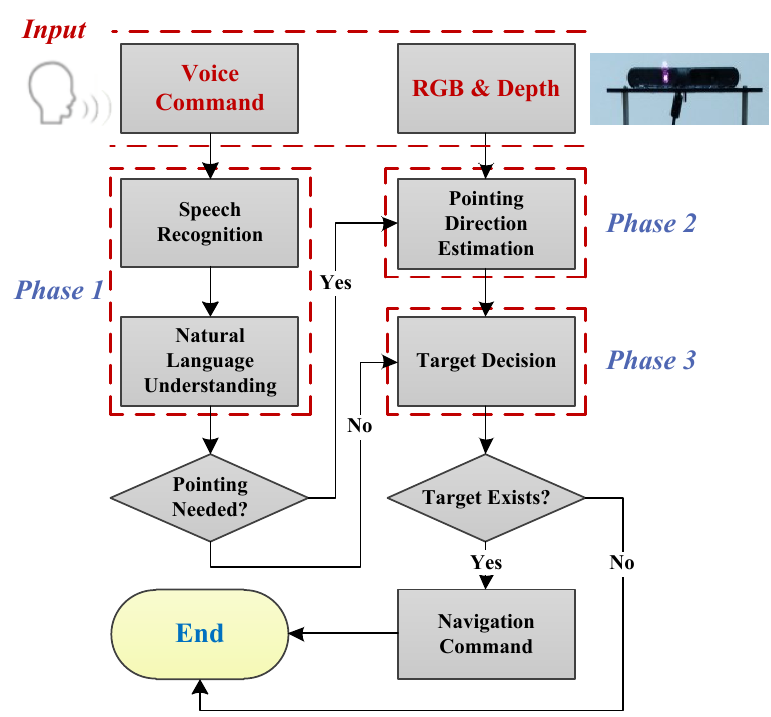}
  \caption{Overall procedure of VGPN. VGPN receives voice command from 
  the user and recognizes speech as natural language text, then natural language 
  understanding is executed. If a pointing gesture is needed, VGPN estimates the pointing 
  direction of the user, otherwise it directly decides the target position. If 
 the navigation target exists, VGPN navigates the robot to the target position.}
  \label{overview}
\end{figure}

\subsection{Phase 1: Voice Understanding} 

In this phase, we perform voice understanding. It consists of two steps, namely 
speech recognition and natural language understanding. 

Firstly, the robot is awakened by a wake-up word, and once awake it receives a voice 
command from the user. Then VGPN uses a Speech-to-Text (STT) 
engine to recognize the voice command as natural language text. Secondly, VGPN translates the natural language text into an intermediate 
instruction, which can be understood by the robot.

In the second step, we perform the following actions: 
\begin{enumerate}[(1)]
\item Parse the natural language text into a dependency model.
\item Parse the dependency model into a unique and equivalent 
string-representation. 
\item Map the string-representation to an instruction template, which is a 
structured intermediate representation.
\item Replace corresponding markers in the instruction template with words from the natural language text.
\end{enumerate}

In the implementation, we use LTP (Language Technology Platform) 
\cite{Che2010LTP} dependency parser to parse Chinese commands into a dependency 
model. Supporting English commands is  future work. Figure 2 shows how the 
sentence ``go to that chair'' in Chinese is parsed into a intermediate 
instruction ``goto (chair, that)''. 

\begin{figure}[thpb]
  \centering
  \includegraphics[width=0.70\linewidth]{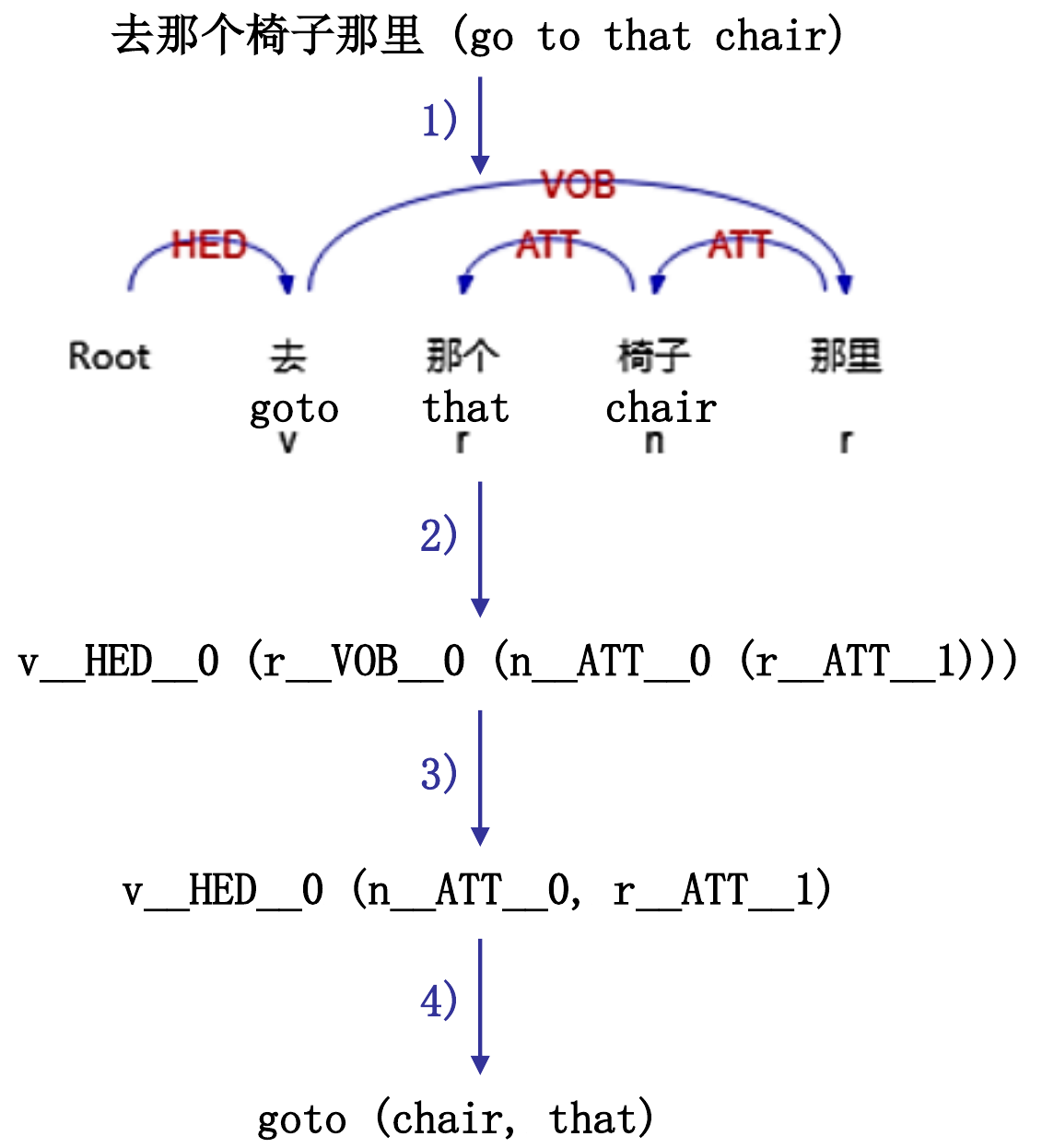}
  \caption{Understanding process of the sentence ``go to that chair''. 1) Parse 
  ``go to that chair'' into a dependency model. 2) Parse the dependency model into 
  a unique string-representation ``v\_\_HED\_\_0 (r\_\_VOB\_\_0 (n\_\_ATT\_\_0 
  (r\_\_ATT\_\_1)))''. 3) The string-representation corresponds to 
  the instruction template ``v\_\_HED\_\_0 (n\_\_ATT\_\_0, r\_\_ATT\_\_1)''. 4) 
  Replace ``v\_\_HED\_\_0'' with ``goto'', replace ``n\_\_ATT\_\_0'' with 
  ``chair'', replace ``r\_\_ATT\_\_1'' with ``that''.
  }\label{paser}
\end{figure}

Similarly, ``go there'' can be parsed into ``goto (there)'', ``go to 
that black chair'' can be parsed into ``goto (chair, black, that)'', and ``turn 90 degree left'' can be parsed into ``turn (left, 90, degree)''. 

If the output instruction indicates that pointing gesture is unnecessary, VGPN will skip 
pointing direction estimation. In fact, there are two common cases in which no pointing gesture is 
 needed: 1) The output instruction does not contain demonstrative pronouns. For example, the user says ``go forward''. 2) The 
output instruction contains a description of the target object and the object is unique in the 
environment. For example, the user says ``go to that door'', and there is only 
one unique door in the environment.

\subsection{Phase 2: Pointing Direction Estimation}

In this phase, we use two keypoints of the human body to generate a vector which 
represents the pointing direction of the user. Specifically, we use the vector 
from the user's eye to the user's wrist as the pointing direction. This 
vector has been proved to achieve better accuracy 
compared to other vectors. Nickel et al. 
\cite{Kai2007Visual} find that people tend to look at the pointing target when 
they perform a pointing gesture, and Abidi et al. \cite{Abidi2013Human} find 
that 62\% of participants were satisfied with the pointing direction from eye 
to hand/finger. 

To estimate the pointing direction, after depth registration, by using rgb and depth data, we 
can get the 3D coordinate of the extracted body keypoints. The pointing direction 
can be formed as the equation:
\begin{equation}
\vec{D} = T_{c\rightarrow m}(\vec{J}_{e} + t(\vec{J}_{w} - \vec{J}_{e}))
\end{equation}
where $\vec{D} \in \mathbb{R}^3$ represents the parameterized pointing direction; 
$T_{c\rightarrow m}$ represents the transformation from camera frame $c$ to map 
frame $m$; $\vec{J_e} \in \mathbb{R}^3$ and $\vec{J_w} \in \mathbb{R}^3$ represent the 3D eye and wrist coordinate in camera frame, respectively; 
$t\in{\mathbb{R}^{+}}$ represents a non-negative real-valued parameter of the 
parameterized pointing direction.

In our implementation, we use the open source library OpenPose\cite{cao2017realtime,simon2017hand,wei2016cpm} to detect human body 
keypoints. OpenPose provides state-of-the-art approach of 2D real-time multi-person keypoint detection, which enables the robot to recognize pointing direction in a multi-person scenario. 

Figure 3 shows the result of pointing direction estimation. We use two 
pointing direction configurations, which are the vectors from right eye to 
right wrist (REW) and from left eye to left wrist (LEW). If the angle between REW and the main body part (which is assumed to be vertically downwards) is 
lager than that between LEW and the main body part (see Figure 3(a1)), the 
pointing direction will be REW (see Figure 3(a2)). Otherwise (see Figure 
3(b1)), the pointing direction will be LEW (see Figure 3(b2)).

We also handle some exceptional cases in this phase. For example, when 
there are no people in the image, the robot will say ``Sorry, I can't see 
you!'' to notify the user, by applying a Text-To-Speech (TTS) service; 
when the user does not perform a pointing gesture, the robot will say 
``Sorry, where are you pointing at?''.

\begin{figure}[thpb]
  \centering
  \includegraphics[width=0.95\linewidth]{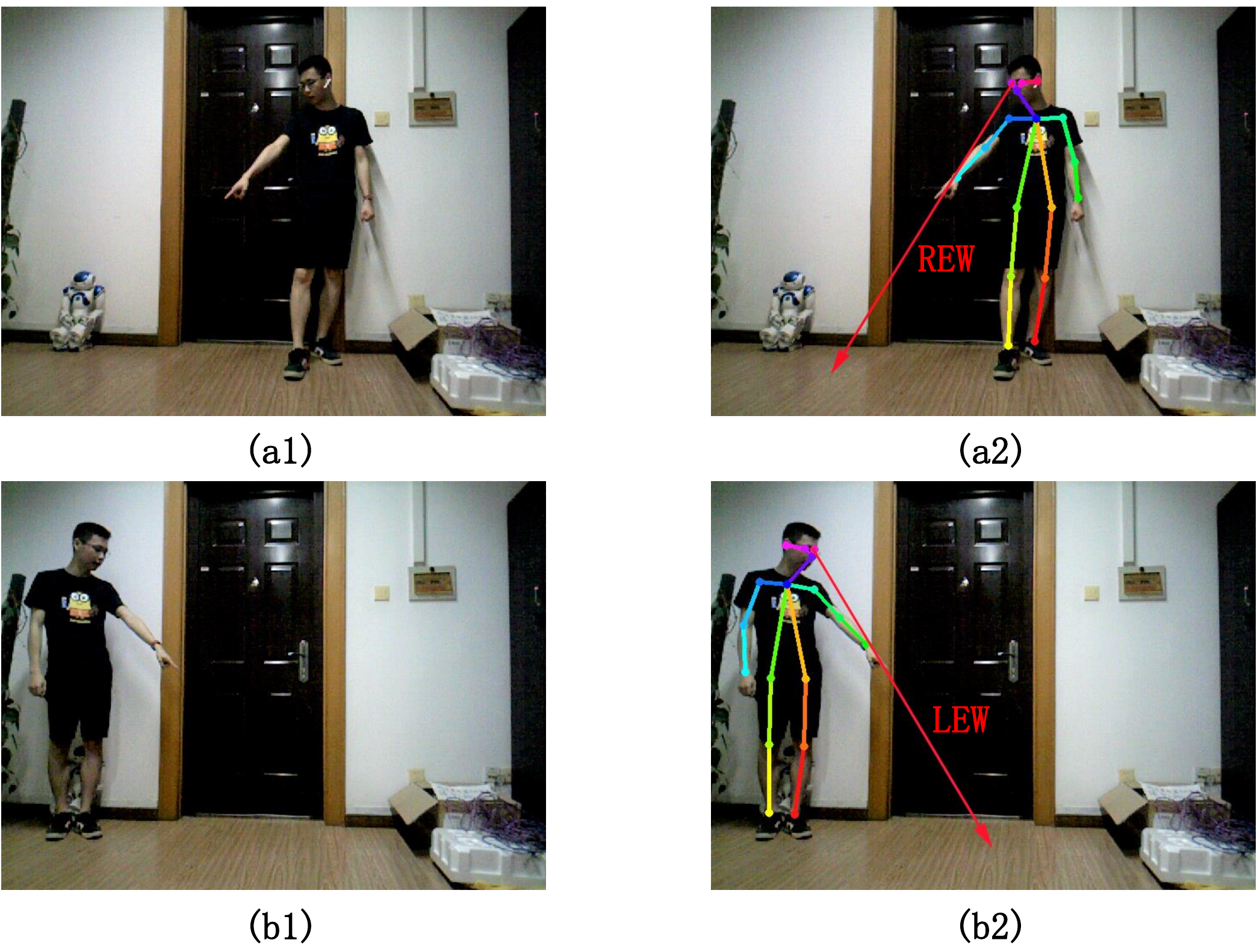}
  \caption{The result of pointing direction estimation. (a2) shows the REW vector of (a1) and (b2) shows the LEW vector of (b1).}
  \label{openpose}
\end{figure}

\subsection{Phase 3: Target Decision}
In this phase, we decide the target position according to the output 
instruction generated in phase 1 and the pointing direction estimated in phase 
2.

When the output instruction contains a description of the target object, for example 
``goto (chair, black, that)'', the navigation target is regarded as the 
position of the described object. Otherwise, for example, ``goto (there)'', the 
navigation target is regarded as the intersection point between the pointing 
direction vector and the ground.

However, errors coming from some aspects may decrease the accuracy of pointing 
navigation, mainly including:1) the user's pointing direction is inaccurate; 2) the 2D coordinate of the body joints extracted by the OpenPose is inaccurate; 3) the depth data from the RGB-D camera is inaccurate, so the 3D joints coordinate in camera frame is inaccurate; 4)the transformation between camera and robot center is inaccurate; 5)the robot position estimated by the SLAM approach is inaccurate.

In order to improve the accuracy of determining the indicated object, combining voice information is a practical solution \cite{Mizuno2003Informing, Sugiyama2005Three}. If the voice command contains a description of the 
target object, we can decide the target location with the following steps: 
\begin{enumerate}
\item Firstly, we calculate the intersection point $P$ between the pointing 
direction vector and the ground.
\item Secondly, we add the object in the environment that satisfies the description into a candidate target set $T$.
\item Thirdly, we calculate the Euclidean distance between the intersection point $P$ and the objects in set $T$.
\item Finally, we identify the location of the object in set $T$ that has the smallest Euclidean 
distance as the navigation target.
\end{enumerate}

\section{EVALUATION}
In this section, we evaluate the prototype of VGPN on a real robot to validate its efficiency 
and navigation accuracy, compared to the pointing-only approach.

\subsection{Experimental Setup}
The evaluation runs on the TurtleBot2 platform, as shown in Figure 4. 
The robot is equipped with an Asus Xtion Pro Live RGB-D camera and a Sick 
TIM561 2D range sensor. The laptop is 
equipped with an Intel core i7-8750H (6 cores @2.2GHz) CPU and a NVIDIA GeForce GTX 1080 GPU. The operating system of the laptop is Ubuntu 14.04 64-bit. The mobile base is an iClebo Kobuki, and the height of the robot is about 
67.5 centimeter. 

\begin{figure}[thpb]
  \centering
  \includegraphics[scale=0.35]{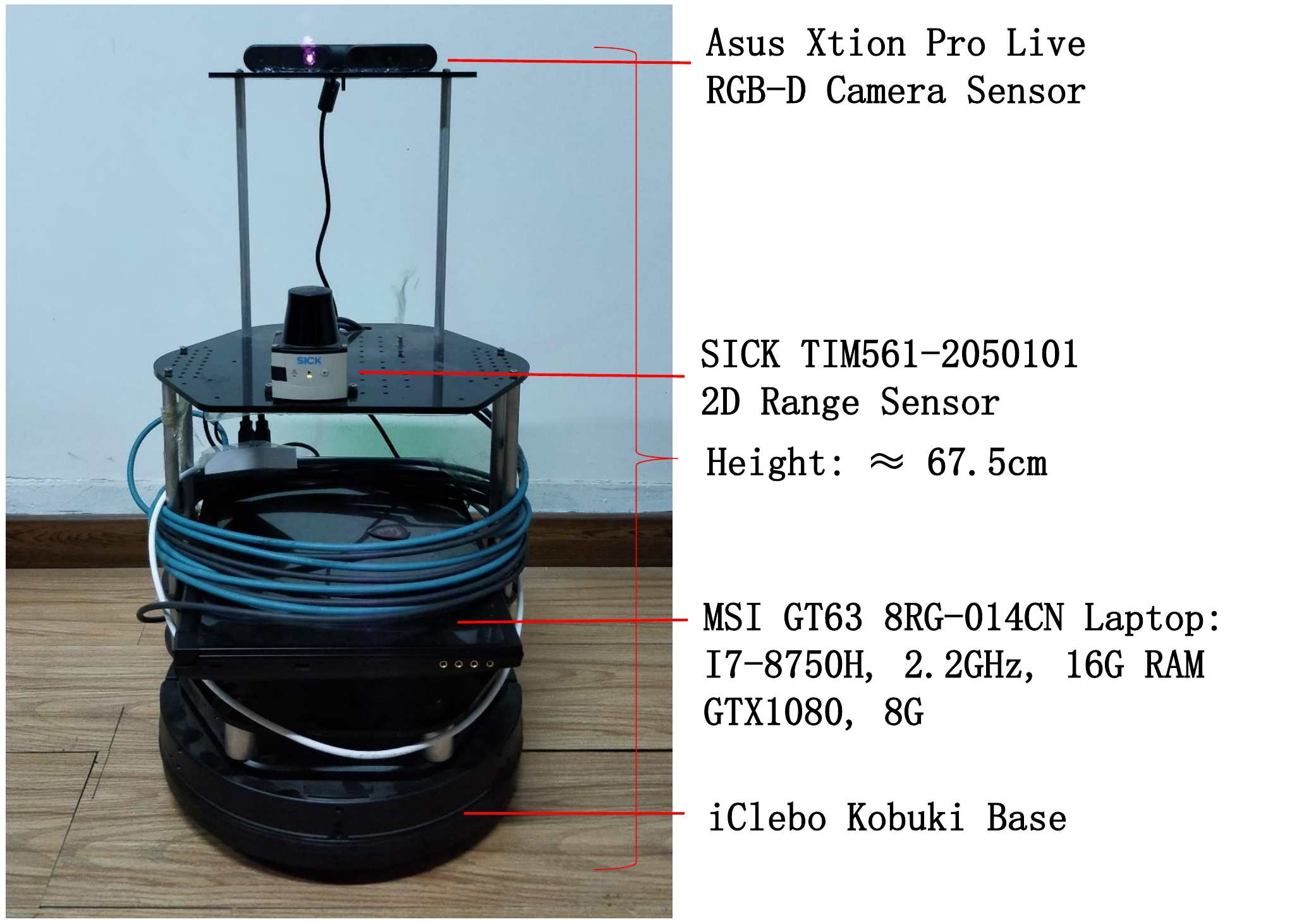}
  \caption{The robot platform. TurtleBot 2 platform equipped with 
  a RGB-D camera, a 2D range sensor and a Kobuki mobile base. }
  \label{platform}
\end{figure}

The VGPN prototype is implemented on ROS (Robot Operating System). This prototype consists of five parts: 
1) a speech recognizer\footnote{https://github.com/cmusphinx/pocketsphinx/\label{pocketsphinx}}; 2) a speech synthesis module\footnote{https://data.aliyun.com/product/nls/\label{ali}} ; 3) VGPN core implementation; 4) a 
Cartographer \cite{Hess2016Real} 2D lidar SLAM module; 5) a path planner and movement control module\footnote{http://wiki.ros.org/move\_base/\label{move_base}}.

We assume that the robot has already built a navigation map by applying the SLAM approach. To give the robot  knowledge about the 
environment, the position coordinate and property of objects have been annotated beforehand. We use the PocketSphinx\textsuperscript{\ref {pocketsphinx}} STT engine for 
recognizing speech as natural language text, and we apply the aliyun\textsuperscript{\ref {ali}} TTS service to create speech from text. Alternative speech recognizers and speech synthesis modules can  be easily applied to our system. 

Figure 5 shows a typical interaction scenario. The user wears a Bluetooth 
headset, then points at a target place and issues a voice command. VGPN understands the voice command and 
estimates pointing direction from rgb (see Figure 5(a)) and depth (see 
Figure 5(b)) data. After determining the target position by combining 
the voice command and the pointing direction, VGPN publishes a navigation goal (see Figure 5(c)). Finally, a proper path is planned\textsuperscript{\ref {move_base}}, and the 
robot navigates itself to the target position. 

\begin{figure}[thpb]
  \centering
  \includegraphics[width=0.90\linewidth]{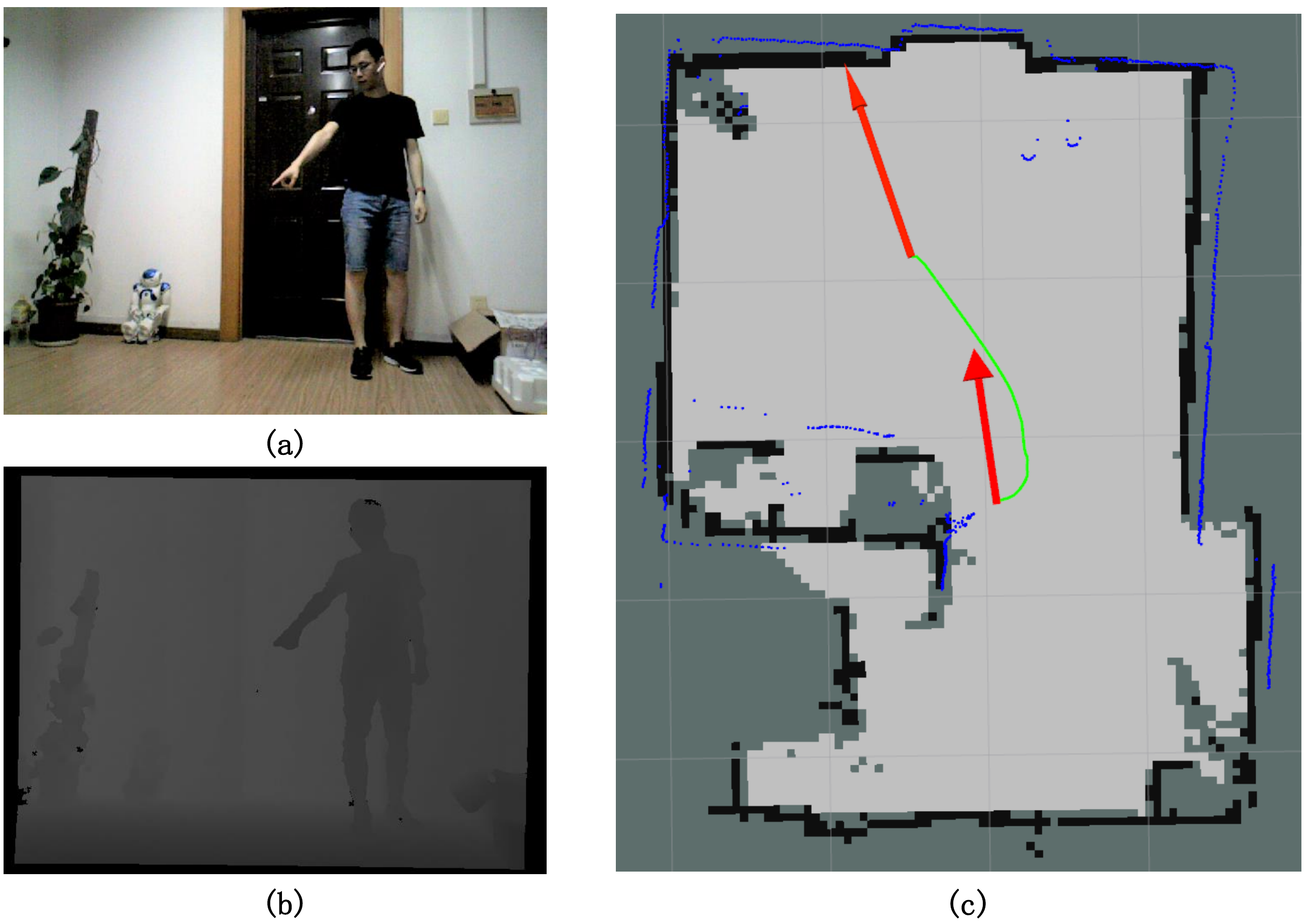}
  \caption{The user interacts with the robot. (a) The received rgb image. (b) The 
  received depth image. (c) The navigation map. The blue dots represent laser 
  scan points. The lower red arrow indicates the current position of the 
  robot, the upper red arrow indicates the target position, the green line 
  indicates the planned path.}
  \label{direc_demo}
\end{figure}

To show the effectiveness of voice guidance, we compared VGPN to the pointing-only approach that without voice guidance. The pointing-only approach recognizes pointing gestures all the time, and treats the 
intersection point between the pointing 
direction vector and the ground as the target position.

\subsection{Efficiency Analysis}
In this section, we validate the efficiency of VGPN compared to the 
pointing-only approach. 

We recruited five volunteers, and asked them to use VGPN and the pointing-only approach 
to navigate the robot. Each volunteer tries 20 times. The volunteer says 
``go to that door" and points at the door that is unique in the environment. We 
measure the processing time cost of the pointing-only approach and VGPN, including: 
\begin{itemize}
\item T1: the time cost of speech recognition.
\item T2: the time cost of pointing direction estimation.
\item T3: apart from T1 and T2, the time cost of all other parts(such as 
message transportation and intersection 
point calculation) from receiving voice 
command to publishing navigation goal.
\item T: total processing time cost, which is the time from receiving voice 
command to publishing navigation goal.
\end{itemize}

\begin{table}[!ht]
\centering
 \caption{The processing time cost of the pointing-only approach and VGPN, including mean and SD 
 (standard deviation)}
 \label{tab:runtime}
 \begin{tabular}{l|c|c} 
  \hline
  Part &  \tabincell{c}{Pointing-only(ms) \\Mean($\pm$SD)} & \tabincell{c}{VGPN(ms) \\Mean($\pm$SD)}\\ 
  \hline 
 T1~ & 0($\pm$0) & 21.38($\pm$8.88) \\ 
 T2~ & 123.80($\pm$9.76) & 0($\pm$0) \\ 
 T3~ & 27.93($\pm$7.08) & 9.23($\pm$2.68) \\ 
 T~ & 151.73($\pm$10.89) & 30.61($\pm$9.40) \\ 
  \hline 
 \end{tabular} 
\end{table}

Table I shows the processing time cost of the pointing-only approach and VGPN. The pointing-only approach only uses voice to trigger the pointing direction estimation, and it does not need to recognize speech, so T1 for the pointing-only approach is 0. VGPN can effectively skip unnecessary pointing direction estimations, so T2 for 
VGPN is 0. Because VGPN does not need to fetch image and calculate intersection 
point in this case, T3 for VGPN is reduced from 27.93ms to 9.23ms in average. Even with a NVIDIA GeForce GTX 1080 GPU, the state-of-the-art approach \cite{cao2017realtime} of multi-person keypoint detection spends about 100ms. Particularly, 
running with CPU is 
$\scriptsize{\sim}$50x slower than that with GPU.

From the results, VGPN in total decreases the processing time cost of pointing robot navigation 
from 151.73ms to 30.61ms, so it reduces about $\frac{151.73-30.61}{151.73}\approx 
79.8\%$ unnecessary processing time cost, when pointing gesture is unnecessary. 

\subsection{Accuracy Analysis}
In this section, we evaluate the accuracy of VGPN compared to the pointing-only 
approach in two aspects:

\textbf{Accuracy of intersection point.}
When the voice command does not contain a description of the target object, for example, the 
command is ``go there'' , the navigation target is the intersection 
point between the pointing direction vector and the ground. To validate the 
accuracy of this case, we performed an experiment as follows.

We collect 125 pointing navigation results from 5 volunteers. Each volunteer 
points at 5 target positions, and each position is pointed at by 5 different 
pointing gestures. Two of the 5 positions are near (distance$\leqslant$2m) from the 
volunteer, two of the 5 positions are middle (2m$<$distance$\leqslant$3m) from the 
volunteer, and one of the 5 positions is far (distance$>$3m) from the volunteer. When 
the volunteer points at different targets, the robot was placed in different 
positions and localized by the SLAM approach.

\begin{table}[!ht]
\centering
 \caption{Intersection point's mean offset and standard deviation (SD) of 
 X-axis, Y-axis and Distance in 125 pointing gestures performed by 5 volunteers.}
 \label{tab:accuracy}
 \begin{tabular}{l|c|c|c|c|c|c} 
  \hline
  \multirow{2}*{Position} & \multicolumn{3}{|c}{Mean offset(cm)} & \multicolumn{3}{|c}{SD(cm)} \\ 
  \cline{2-7} 
 ~ & X  & Y & Dis & X  & Y & Dis \\ 
 \hline 
 near~ & 13.13  & 9.35 & 17.85 & 6.99 & 7.32 & 6.27 \\ 
 middle~ & 22.06  & 15.82 & 30.30 & 12.57 & 12.64 & 11.29 \\ 
 far~ & 41.006 & 68.47 & 82.26 & 34.10 & 36.23 & 43.37 \\ 
  \hline 
 \end{tabular} 
\end{table}

Table II shows the results. We compute the intersection point's mean offset for x-axis, y-axis and distance from the position where the volunteer is supposed to point at, and 
we also compute the intersection point's standard deviation for x-axis, y-axis and distance. 

Because it is more difficult to precisely point at a further position than a 
closer one, the mean accuracy of intersection point decreases as the 
distance increases. 

\textbf{User satisfaction with navigation accuracy.}
When the voice command contains a description of the target object, for example ``go to that black chair'' , the navigation target is the position of the described object. To validate the 
accuracy of this case, we performed the experiment as follows.

We put two objects 20 cm away from each other in the 
environment. Five volunteers stand 2m away from the two objects, and they use the
pointing gesture and a voice command that describes the  target   to navigate 
the robot to one of the two objects. Each volunteer performs 10  
pointing gestures, and the pointing direction is a little inaccurate. For 
example, the volunteer says ``go to that chair'' and points at a position that  is
near to the target chair. Each volunteer uses the pointing-only approach and VGPN, 
and reports their satisfaction score for navigation accuracy with a score 1 to 5 (5 is 
the best). 

The satisfaction score qualitatively measures the navigation accuracy, and it is higher when navigation destination of the robot is closer to the target position. For example, when the user says ``go to that chair'' and points at the bed nearby the desired chair, the satisfaction score will be very low if the robot goes to the bed rather than the chair. We conducted an experiment called SAME when the two objects belong to the 
same category and have the same property, and conducted another experiment, DIFF, when the two objects belong 
to different categories.

\begin{figure}[thpb]
  \centering
  \includegraphics[scale=0.60]{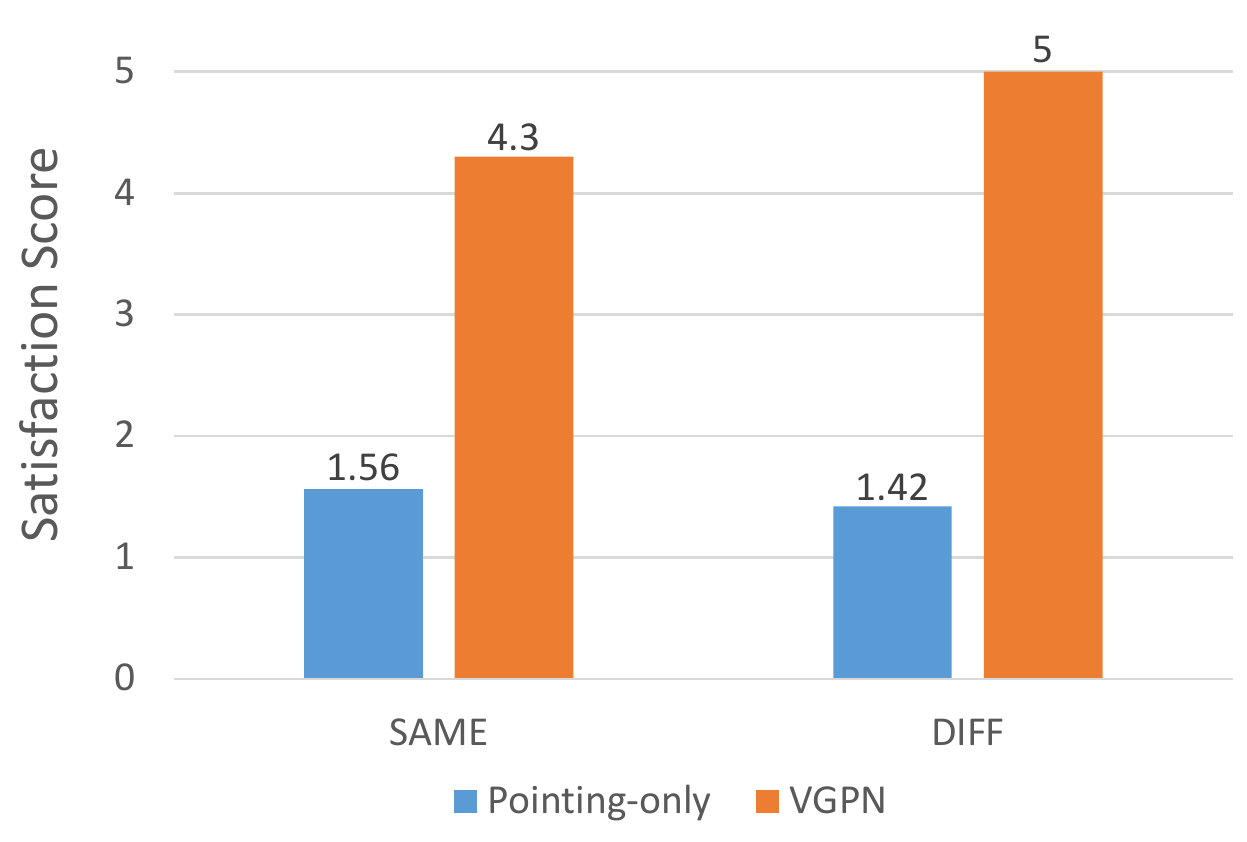}
  \caption{The satisfaction score (5 is the best) of navigation accuracy for 
  the pointing-only approach and VGPN. The experiment SAME is for the two objects belonging 
  to the same category and have the same property, and the experiment DIFF is for the two objects belonging to 
  different categories. }
  \label{accuracy}
\end{figure}

As shown in Figure 6, almost all volunteers were not very satisfied with the 
navigation goal when using the pointing only. But by using VGPN, almost all
volunteers were satisfied with the navigation accuracy. There are some cases that the volunteer points at a position which is nearer to the undesired target. In these cases for experiment SAME, VGPN may fail to navigate the robot to the desired location, and the score is lower than 5.

In the experiment, by using voice guidance in pointing navigation, VGPN can effectively navigate the robot to the desired target place, even though pointing direction of the target is inaccurate. 

\section{CONCLUSION}
In this paper, we propose a voice-guided pointing approach for robot navigation, named VGPN. VGPN consists of three phases, namely voice understanding, pointing 
direction estimation and target decision. It allows the user to use voice and pointing gestures, to navigate the robot to a desired target 
position. VGPN recognizes pointing gestures only if the voice information is not sufficient for robot navigation, and determines target positions by combining voice commands
and pointing directions. We evaluated VGPN on a wheeled robot, TurtleBot 2, and 
compared it to the pointing-only approach. The results show that VGPN can 
reduce 79.8\% processing time cost of pointing robot navigation when pointing gesture is 
unnecessary, and VGPN can also improve the user satisfaction with navigation 
accuracy.

In the future, we plan to replace the Cartographer 2D SLAM with a 3D SLAM 
approach such as RTAB-Map \footnote{http://introlab.github.io/rtabmap/}. Additionally, we plan to support English commands by using 
the Stanford dependency parser\cite{Chen2014A}.

\bibliographystyle{IEEEtran}
\bibliography{IEEEabrv,robio_bib}

\end{document}